\documentclass[10pt,twocolumn,letterpaper]{article}

\usepackage{cvpr}      % To produce the REVIEW version
\usepackage[T1]{fontenc}   % modern, 256-glyph encoding
\usepackage[utf8]{inputenc}% only needed with pdfLaTeX
\usepackage{tabularx}

\usepackage{xcolor}     % before you use \color

\definecolor{cvprblue}{rgb}{0.21,0.49,0.74}
\usepackage[pagebackref,breaklinks,colorlinks,allcolors=cvprblue]{hyperref}
\usepackage{listings}

\title{Understanding Trade-offs When Conditioning Synthetic Data}

\author{%
  Brandon Trabucco\textsuperscript{1} \quad
  Qasim Wani\textsuperscript{2} \quad
  Benjamin Pikus \textsuperscript{2} \\[4pt]
  Vasu Sharma \textsuperscript{3} \\[4pt]
  \textsuperscript{1}Carnegie Mellon University \quad
  \textsuperscript{2}Advex AI\\
  \textsuperscript{2}Facebook AI Research\\
  {\tt\small brandon@btrabucco.com, qasim@advexai.com, ben@advexai.com, vasus@andrew.cmu.edu}
}

\begin{document}
\maketitle
\begin{abstract}
Learning robust object detectors from only a handful of images is a critical challenge in industrial vision systems, where data collection can take months due to the difficulty of acquiring high-quality training data. Synthetic data is becoming a key solution to increase data-efficiency for industrial vision tasks such as visual inspection and pick-and-place robotics. Current solutions utilize 3D modeling tools such as Blender and Unreal Engine to create synthetic data which offer great controllability for the user but take several weeks to generate a handful of images. Additionally, 3D-generated images frequently suffer from a high sim-to-real gap, limiting their effectiveness in practice. Recent advances in Diffusion models 
offer a promising paradigm shift, enabling the production of high-quality images in minutes instead of weeks. However, controlling these models remains challenging, particularly in low-data regimes. While the community has developed numerous adapters to extend diffusion models beyond textual prompts, the impact of different conditioning methods on synthetic data quality remains unclear. In this work, we analyze 80 diverse visual concepts from four standard object detection benchmarks to compare prompt-based and layout-based conditioning. We find that when conditioning diversity is low, prompts produce higher-quality synthetic data, but as diversity increases, layout conditioning becomes superior. When conditions match the full training distribution, layout-conditioned synthetic data improves mean average precision (mAP) by an average of 34\%—and up to 177\%—compared to real data. Open source code for reproducing our study.

% We conduct an analysis on 80 diverse visual concepts adapted from four standard object detection benchmarks, and compare the impact of prompt conditions versus layout conditions on synthetic data quality. We find a subtle result: when conditions are less diverse, prompts beat layouts for synthetic data quality, but as diversity improves, layouts beat prompts. When the distribution of conditions matches full training distribution, layout-conditioned synthetic data improves mAP (mean average-precision) on average 34\%, up to 177\% versus real data. Open source code for reproducing our study, and generating synthetic data will be available at the following website: \href{https://diffusion-augmentation.github.io}{diffusion-augmentation.github.io}.
\end{abstract}
\section{Introduction}

Learning from a handful of example images is one of the most important challenges in vision, and crucial for industrial tasks, where data collection and labeling can require specialized knowledge, and months of effort from dedicated engineering teams \cite{FordQualityControlExample}. Synthetic data is a promising solution to this data problem, building on the visual fidelity of generations from Diffusion Models \cite{StableDiffusion,SDXL,Imagen,ramesh2021zero,dai2023emu,unCLIP}. As generation capabilities broaden, and images become more realistic, a new paradigm is emerging, using generative models to enhance real data \cite{DAFusion,DatasetDM,InstanceAugmentation}. Starting from a handful of real images, Diffusion Models can generate high-quality examples for bespoke concepts present in data \cite{DAFusion,InstanceAugmentation}, but naive sampling can produce synthetic data that is less valuable than an equivalent amount of real data \cite{Azizi2023SyntheticDF}. To understand the design space for synthetic data, and find when synthetic examples are most valuable, we isolate conditioning as a primary factor to analyze. The community has trained numerous adapters for Diffusion Models that extend their conditions beyond simple textual prompts \cite{LoRA, ControlNet}, including canny edge maps \cite{CannyEdgePaper}, drawings, and depth images. How do conditions impact the quality of synthetic data?

\begin{figure*}
    \centering
    \includegraphics[width=\linewidth]{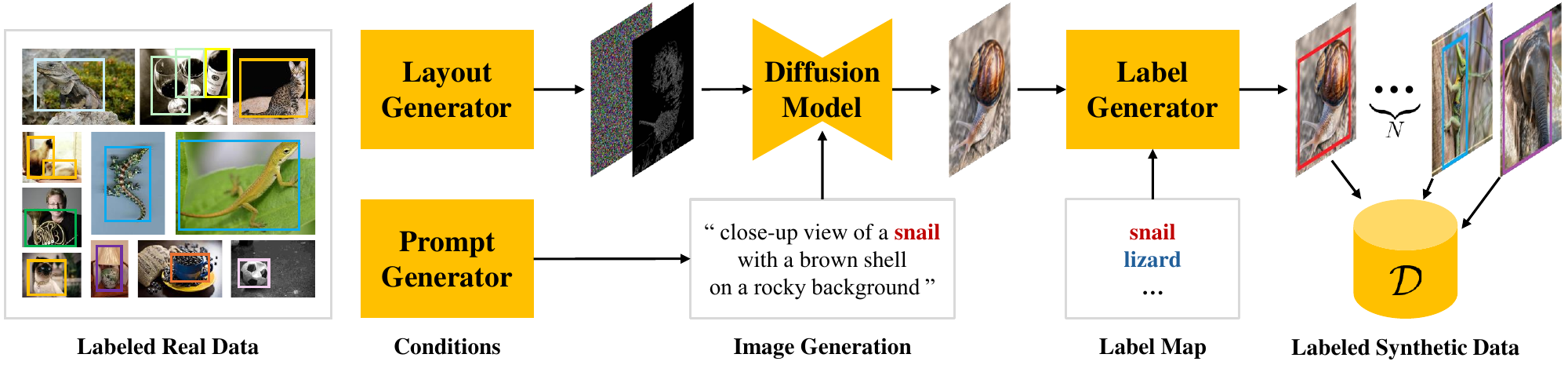}
    \caption{\textbf{Workflow for generating labeled synthetic data for object detection.} Starting from a collection of real images labeled with instance bounding boxes, we explore the impact of the type of conditioning (prompts and canny edge maps), and the source of donor images that conditions are drawn from (incrementally increasing the size of the donor image pool). Our ablation highlights a key trade-off: with relatively few donor images, conditioning via prompts alone produces the best synthetic data, but as the number of donor images increases, it becomes optimal to employ more conditions drawn from the donor images.}
    \label{fig:main-experiment-workflow}
\end{figure*}

We analyze the impact of different types of conditions on the quality of synthetic data generated by Diffusion Models across four standard object detection datasets. We select 80 diverse visual concepts from these datasets, and compare prompt conditions, the most popular type of conditioning used in current works \cite{DAFusion,RealGuidance,Azizi2023SyntheticDF,DatasetDM}, to layout conditions. These types of conditions represent two extremes---prompts are effective high-level guidance \cite{StableDiffusion,SDXL,Imagen}, but rely on the visual prior learned by Diffusion Models to generate realistic layouts, which are non-trivial to control via prompts. In contrast, layouts enable precise spatial control, but can hinder diversity of generations if poor layouts are selected. Previous works have employed prompt-based \cite{Azizi2023SyntheticDF,DatasetDM}, layout-based \cite{InstanceAugmentation}, and hybrid  \cite{DAFusion,RealGuidance} conditions, but the trade-offs resulting from the granularity of conditions, independent from the method, are relatively unexplored. \textit{We aim to design the simplest experiment to uncover these trade-offs, using existing models and adapters.} Using SDXL \cite{SDXL}, we employ a pretrained ControlNet \cite{ControlNet} from Huggingface, and explore the impact of the conditions provided to the diffusion model on the performance of few-shot object detection models trained on real data and synthetic data, via the faithfulness of conditions to the real data distribution.

We uncover a subtle result: \textit{conditioning on just prompts is best when a relatively weak source of conditions is provided, but as the faithfulness of conditions to the real distribution improves, adding layouts begins to produce significantly better synthetic data than prompts alone.} In a data-limited setting, layout-conditioned synthetic data improves mAP by 34\% on average, and up to 177\% versus real data, when layouts are faithful to the real distribution. While Diffusion models have a strong visual prior, this result implies that techniques for analyzing the distribution of real data, and generating faithful layouts are likely to produce significant gains for synthetic data on vision tasks. Open source code for reproducing the experiments in our study, and generating synthetic data given layouts will be available at the following website: \href{https://diffusion-augmentation.github.io}{diffusion-augmentation.github.io}.
\section{Related Works}

\paragraph{Few-Shot Learning}

Few-shot learning in computer vision focuses on recognizing novel object categories with limited annotated examples. Few-shot object detection (FSOD) often employs fine-tuning strategies, wherein large pre-trained models such as OwlV2 \cite{Minderer2023ScalingOO} or SAM \cite{SegmentAnything} are adapted using real labeled examples from the target domain.

Alternatively, meta-learning approaches pre-train a network on a set of base classes and later fine-tune it to adapt quickly to novel classes \cite{sun2021fsce, wu2020meta, han2022meta, han2022multi, park2022hierarchical}. These methods often involve techniques such as episodic training which simulates few-shot scenarios during learning, and data weighting strategies that emphasize the importance of specific samples. While these techniques can be highly effective within specific domains, their domain-specific nature and reliance on tailored training strategies limit their generalizability to broader applications \cite{jamal2019task, rajeswaran2019meta, kirkpatrick2017overcoming, santoro2016meta}.

A key challenge in FSOD lies in the distribution shift between pre-training and target domains. When this disparity is high, models may require more data to generalize effectively \cite{zoph2020rethinking, hammoud2024pretraining}. In scenarios where collecting real-world examples is prohibitively expensive, synthetic data can offer a practical solution, enriching the target domain with diverse and controlled examples.

\paragraph{Using diffusion for synthetic data generation }

Over the last few years, advent of large-scale diffusion models have shown very impressive results in various tasks such as image generation (Dall-E \cite{ramesh2021zero}, Imagen \cite{saharia2022photorealistic}, and Emu \cite{dai2023emu}) and image editing \cite{wang2024magic, geng2024instructdiffusion} using just text prompts. There have been other works that study the impact of controlled image generation when adding additional inputs beyond text \cite{li2023gligen, kang2024eta}, \cite{patashnik2023localizing, meng2021sdedit}. One such popular technique is ControlNet \cite{ControlNet, li2025controlnet} which is conditioned on several inputs such as text, canny edge \cite{canny}, and segmentation masks to produce more fine-granular details during the image generation process. In our paper, we explore the impact of downstream model performance as a function of adding additional control to the diffusion process which we define as prompt and layout condition.

Several studies \cite{Gen2Det, Rosie, Azizi2023SyntheticDF} have investigated how to use diffusion models to enhance performance of object detection models for supervised tasks such as Yolo \cite{YOLO, YOLOv2, YOLOv3, YOLOv8}, Dino \cite{Zhang2022DINODW, GroundingDINO, oquab2023dinov2}, and MaskRCNN \cite{MaskRCNN}. Unlike classification, the process of generating the label can be tricky. Approaches like InstaGen \cite{Feng2024InstaGenEO}, Gen2Det \cite{Gen2Det} and several others \cite{deshpande2024auto, segmentationdiffusion, saragih2024using} employ the strategy of generating raw images first and then using an off-the-shelf labeler like OwlV2 \cite{Minderer2023ScalingOO} or SAM \cite{SegmentAnything} etc to generate the labels. The challenge with this approach is the labeler becomes the bottleneck in that if the zero-shot performance of the labeler is poor, it will hinder downstream model performance \cite{ma2022effect, chachula2022combating}. To combat this, other papers have studied the problem from acquiring the label first and then generating the images conditioned on these labels \cite{ControlNet, InstanceAugmentation}. In our work, we show how adding more diverse control beyond just text prompt helps improve model performance.
\section{Synthetic Data via Diffusion Models}
\label{sec:background}

For data-limited tasks, synthetic data aims to produce realistic training examples that can serve as a proxy for real samples that may be challenging to collect due to time and cost. Given a distribution of real data $p(x_{\text{real}}, y)$, we will employ a Diffusion Model to produce an image $x_0$ that is likely under the real distribution. Diffusion Models \cite{DDPM,DDIM,StableDiffusion} are a class of sequential latent variable models inspired by thermodynamics \cite{OriginalDiffusionPaper} that produce samples via a Markov chain with learned Gaussian transitions. The chain starts from an initial noise distribution $p(x_T) = \mathcal{N}(x_T; 0, I)$, and iteratively samples latent variables $x_{t}$ with gradually reducing variance that approach the data distribution at the final step.
\begin{gather}
    p_{\theta}(x_{0:T}) = p(x_T) \prod_{t = 1}^{T} p_{\theta} (x_{t - 1} | x_t)
\end{gather}
The transition distributions $p_{\theta} (x_{t - 1} | x_t)$ are designed to gradually reduce variance according to a predetermined schedule $\beta_1, \hdots, \beta_T$ so that, by the final step of the chain, $x_0$ represents a sample from the target data distribution. The choice of variance schedule depends on the underlying diffusion model, and a fixed covariance $\Sigma_t = \beta_t I$ and a learned mean $\mu_{\theta} (x_t, t, c)$ are often used to parameterize $p_{\theta}$. 
\begin{gather}\label{eqn:reverse-step}
    \mu_{\theta} ( x_t, t, c ) = \frac{1}{\sqrt{\alpha_t}} \left( x_t - \frac{\beta_t}{\sqrt{1 - \tilde{\alpha}_t}} \epsilon_{\theta} ( x_t, t, c ) \right)
\end{gather}
This choice of parameterization results from deriving the optimal reverse process \cite{DDPM} that minimizes a variational lower bound on the likelihood of the data, assuming a forward process where each step is a Gaussian distribution. The function $\epsilon_{\theta}(x_{t}, t, c)$ is a neural network that predicts the noise added to a sample at timestep $t$ of the diffusion process, conditioned on $c$, typically a text-based prompt.

\begin{figure*}
    \centering
    \includegraphics[width=1\linewidth]{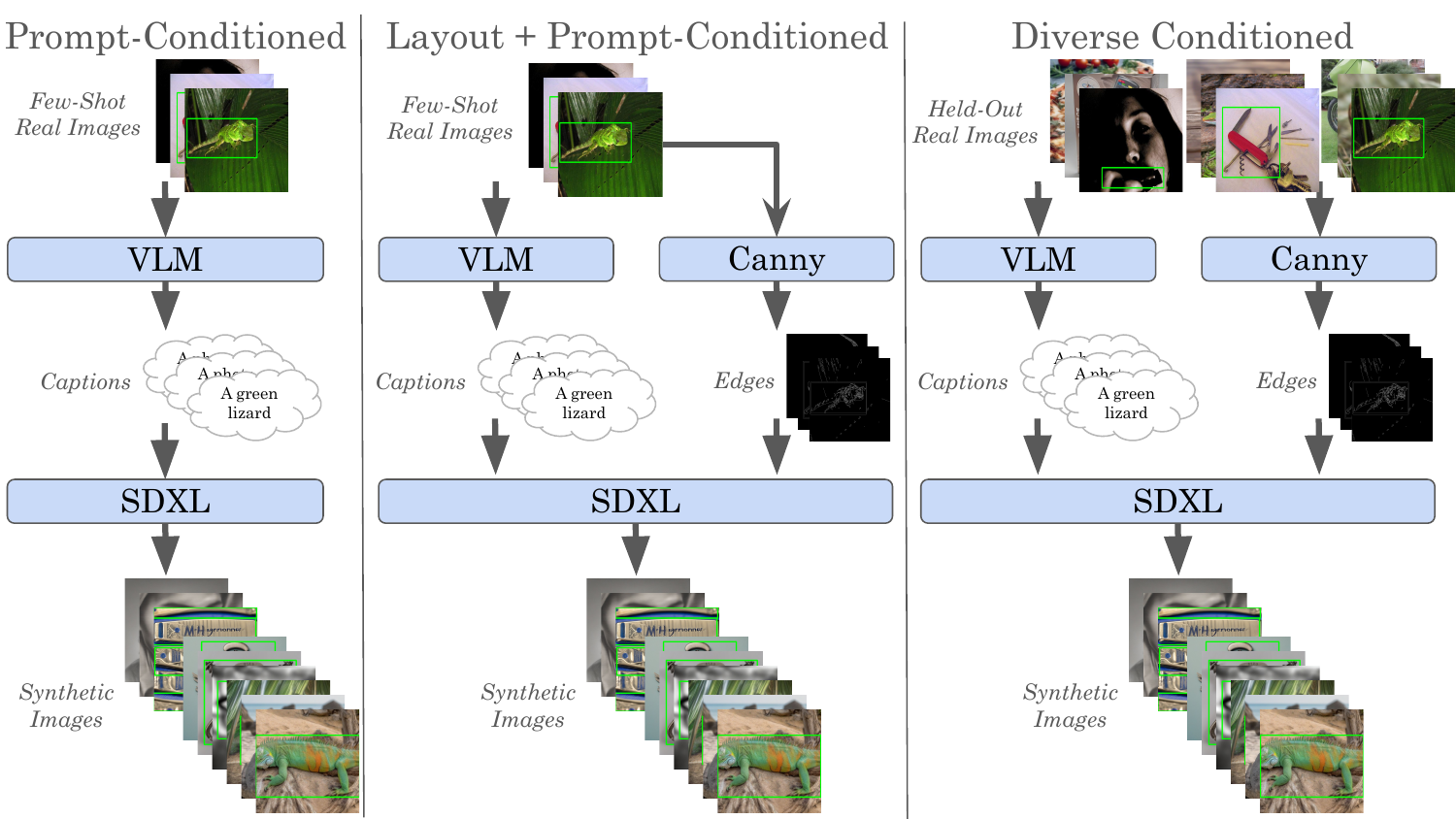}
    \caption{We study three synthetic data strategies. Prompt-Conditioning (left) uses captions from real examples, via a VLM, to generate synthetic examples. Layout+Prompt Conditioning (middle) generates synthetics both with these captions and canny edges from the same real set. Diverse Layout+Prompt Conditioning (right) takes captions and canny edges from a larger, held out set that is not included in downstream training. This held out set represents a more diverse conditioning distribution.}
    \label{fig:pipeline-overview}
\end{figure*}

% Brandon: this section needs a significant re-write. Todo

\paragraph{Types Of Conditions} When generating images for synthetic data, the conditions provided to the Diffusion Models are the central way to steer data quality. Work in prompt-engineering has shown that well-tuned prompts can dramatically improve image quality for Diffusion Models \cite{PromptEngineeringDiffusion}. Understanding the impact of well-chosen conditions (especially those beyond text), is crucial towards generating high-quality synthetic data for domains that are further from the training distribution of Diffusion Models---where their visual prior is weaker.

% Brandon: below is a somewhat revised version of the original section

The second component of synthetic data generation is to produce realistic labels $y$ alongside generated images. Label generation can be challenging because it is intrinsically tied to generation, and requires precise knowledge about the content of generated images. When generating synthetic data, we have the choice to factorize the data distribution \textit{image-first} or \textit{labels-first}. When generating in the image-first direction, which corresponds to the following factorization of the synthetic data distribution, we allow the diffusion model to generate freeform images, and assign labels using a secondary expert labeler.
\begin{equation}
    p(x_{0}, y) = \text{DiffusionModel} (x_{0}) \cdot \text{Labeler} (y | x_{0})
\end{equation}
This choice of data factorization can have important advantages---the principle benefit being that we can use off-the-shelf models for each component. However, the core limitation of the image-first factorization is that we are distilling skills from the expert labeler on generated data, and are fundamentally restricted by the expert's performance. The alternative labels-first approach promises to remove this restriction by conditioning the diffusion model on $y$.
\begin{equation}
    p(x_{0}, y) = \text{Generator} (y) \cdot \text{DiffusionModel} (x_{0} | y)
\end{equation}
Synthetic data in labels-first order is more challenging to generate because it requires a specialized diffusion model, and effective methods for labels-first generation of synthetic data have not yet been developed. We do not aim to solve labels-first generation in this work---rather, we seek to understand how the \textit{optimal labels-first} method compares to the \textit{optimal image-first} method, to identify which strategy one should prioritize as the field matures. To this end, we employ layout conditions (see Section~\ref{sec:image-generation}) as a proxy for conditioning the diffusion model on bounding boxes, and we simulate an expert label generator by taking these layout conditions from real images (see Section~\ref{sec:diversity-is-key}).

\section{Synthetic Data Generation}
\label{sec:synthetic-generation}

%Brandon: TODO, we need a figure showing the entire pipeline.

To facilitate our analysis of the importance of diversity, factorization, and scale, we develop a pipeline for the generation of synthetic images and labels for detection. %Shown in Figure~\ref{fig:pipeline-overview}, 
Our pipeline has three stages: (1) image generation, (2) label generation, and (3) evaluation, where we train detection models on a mix of real and synthetic examples. 

\subsection{Image Generation}
\label{sec:image-generation}

\begin{figure}[ht]
    \centering
    % First PDF
    \includegraphics[width=0.9\linewidth]{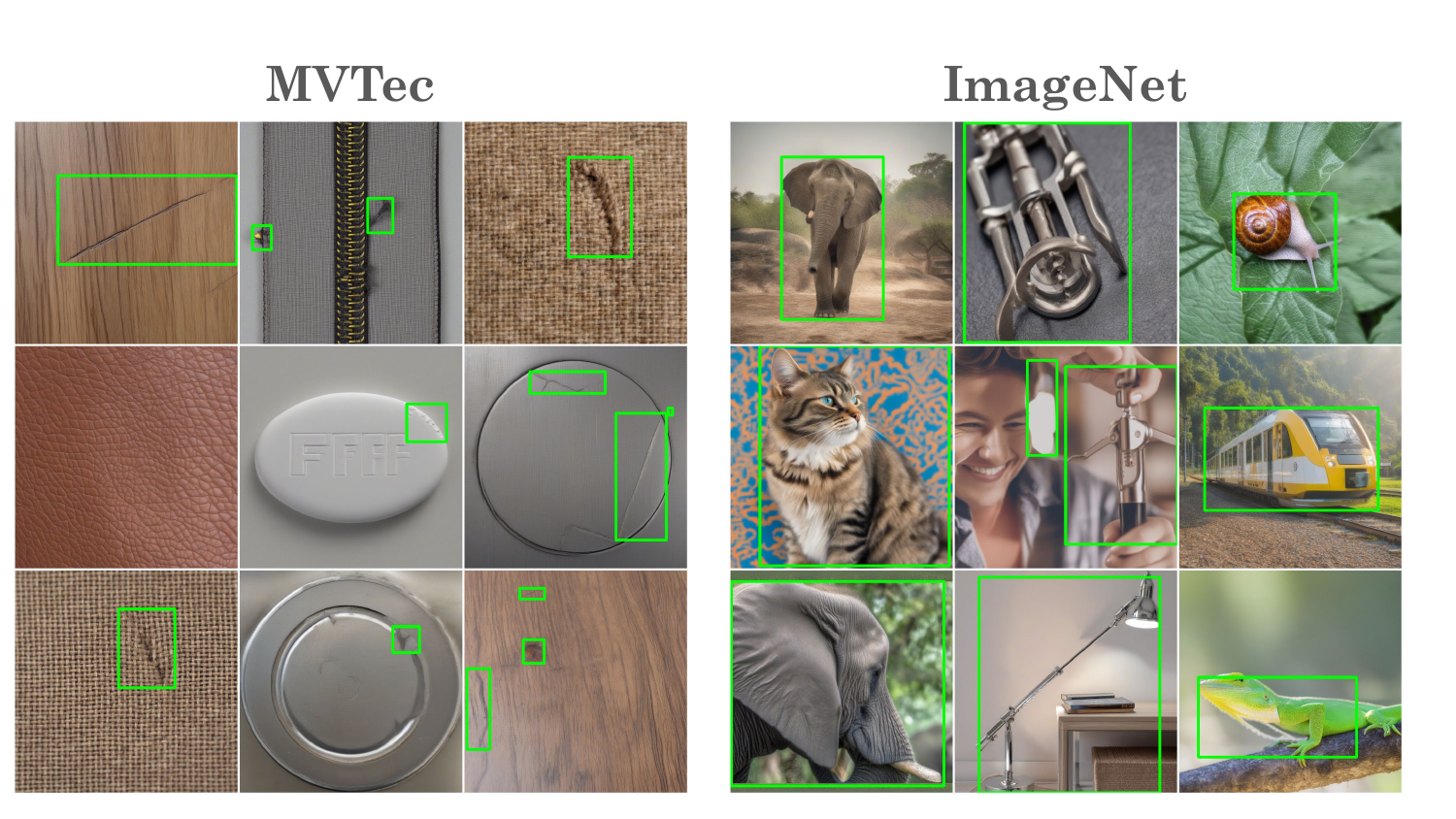}
    \vspace{0.5em} % Adjust vertical spacing if needed
    % Second PDF
    \includegraphics[width=0.9\linewidth]{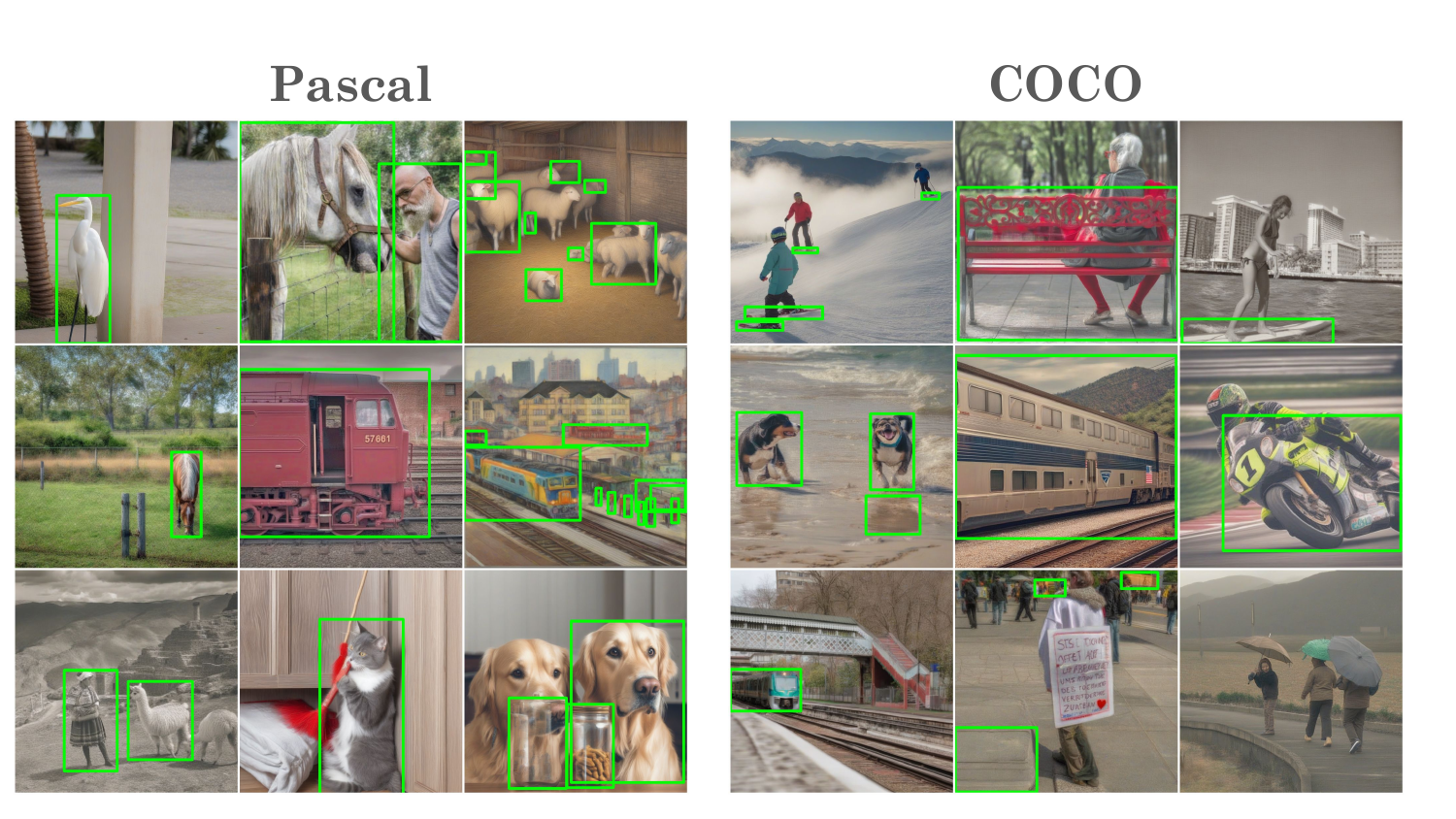}
    \caption{Example generated images and labels using diverse conditions, for ImageNet, MVTec, Pascal, and COCO.}
    \label{fig:pipeline-examples}
\end{figure}

To begin the pipeline, we generate images using SDXL, a latent diffusion model \cite{StableDiffusion,OriginalDiffusionPaper}, that employs an iterative denoising procedure \cite{DDPM,DDIM} to sample images $x_{0} \in \mathcal{R}^{H \times W \times 3}$ starting from an initial distribution $x_{T} \sim \mathcal{N}(0, \Sigma)$, typically a unit Gaussian. Importantly, the diffusion process has several key arguments relevant to our analysis. First, we will study a prompt condition $c_{\text{prompt}} ( x_{\text{real}}, y )$ that embeds a description of a training example in natural language. To extract these descriptions, we employ an off-the-shelf VLM \cite{liu2023llava,liu2023improvedllava,gpt4o} to produce a detailed caption given an image, and a description of the application domain---the exact prompts for each dataset can be viewed in Supplementary ~\ref{app:prompts}. Second, we will study a layout condition $c_{\text{layout}} ( x_{\text{real}}, y )$ that embeds the spatial structure of a training example as an input to the diffusion process. Different representations for the layout condition are possible, and we choose an open source ControlNet \citep{ControlNet} checkpoint based on canny edges, as this is a popular choice for layout control. Equipped with prompt and layout conditions, we steer the diffusion process with these to produce images with precise control using classifier free guidance \cite{ClassifierGuidance}. For experiments using only prompt conditions, we employ standard guidance, and for experiments using both prompt and layout conditions, we employ a fused guidance.
\begin{multline}\label{eqn:image-generation}
    \tilde{\epsilon}_{\theta} \left( x_{t}, t, c_{\text{prompt}}, c_{\text{layout}} \right) = \\ (1 + \lambda) \epsilon_{\theta} \left( x_{t}, t, c_{\text{prompt}}, c_{\text{layout}} \right) - \lambda \epsilon_{\theta} \left( x_{t}, t, \emptyset, \emptyset \right)
\end{multline}
%Generations for various guidance scales are shown in Figure~\ref{fig:pipeline-examples} for prompt-only, and layout+prompt conditioning, and demonstrate how synthetic data can faithfully adhere to layout conditions, exposing a powerful mechanism to control the synthetic data distribution. Equipped with these tools, we consider how to assign effective labels to generations.
Generations using layout + prompt conditioning are shown in Figure~\ref{fig:pipeline-examples} and demonstrate how synthetic data can faithfully adhere to layout conditions, exposing a powerful mechanism to control the synthetic data distribution. Equipped with these tools, we consider how to assign labels to generations.

\subsection{Label Generation}
\label{sec:label-generation}

Synthetic data generation for structured visual tasks like object detection can be challenging because assigning labels requires a granular understanding of the content of images. The goal of this work is to identify the key bottlenecks in the image-generation stage of the pipeline, so we conduct experiments assuming access to an expert detection model. We employ an off-the-shelf detection model Owl-v2 \citep{Minderer2023ScalingOO} for zero-shot label generation, prompted with a label map consisting of the names of each candidate object to detect. While zero-shot labels often vary in quality, we find that optimally mixing real and synthetic data while training, discussed in Section \ref{sec:detection-evaluation}, mitigates problems in label quality. Figure~\ref{fig:pipeline-examples} showcases example images and labels generated by our pipeline, highlighting that images and labels are generally of high-quality, an important requirement for our work. Equipped with tools for generating high-quality images and labels, we can begin to study the key factors that unlock performant synthetic data for object detection tasks. 

\subsection{Evaluating Detection}
\label{sec:detection-evaluation}

%Brandon: currently working on this section.

Following previous work in classification \cite{DAFusion}, we split the training dataset for downstream models into a real part, and a synthetic part that are sampled independently in each batch with a weight controlled by the hyperparameter $\alpha$. For a batch $B$ of $N$ datapoints, each sample in the batch $(x_{i}, y_{i}) = B_{i}$ is real with probability $(1 - \alpha)$, and synthetic otherwise. Crucially, the hyperparameter $\alpha$ allows us to control the mixing ratio of real and synthetic data invariant to the size of each pool of data. This enables training on significantly more synthetic examples than real, without down-weighting the real data in the loss function. We then independently control the number of synthetic examples per real via a second hyperparameter $M$, and set $\alpha=0.5$, $M=10$, a standard practice in prior work for synthetic data \cite{DAFusion}. Discussed further in Section~\ref{sec:results}, these choices correspond to an equal mix of synthetic and real data, and as the quality of synthetic data improves, more aggressive mixing ratios closer to $0.0$ should be used. Regarding $M$, we observe in Section~\ref{sec:results} that performance improves monotonically with synthetic data scale for a fixed data mix, and $M=10$ balances data scale and computational efficiency.

%%Brandon: next paragraph to discuss the object detection model training methodology.

\paragraph{Training Detection Models} Given a mix of real and synthetic data, we evaluate data generation strategies by training detection models with varying amounts of real data. We adapt four standard detection tasks from relevant literature~\cite{Imagenet,COCO,PascalVOC,MVTecAD}, and uniformly sample 32, 64, 128, 256, and 512 real examples for training. This sub-sampling procedure allows studying the viability of synthetic data across multiple data scales. Results in Section~\ref{sec:results} show that synthetic data can significantly outperform traditional augmentation for detection at \textit{both} high and low real data scales.

\section{Key Factors For Synthetic Data}
\label{sec:results}

\begin{figure*}
    \centering
    \includegraphics[width=\linewidth]{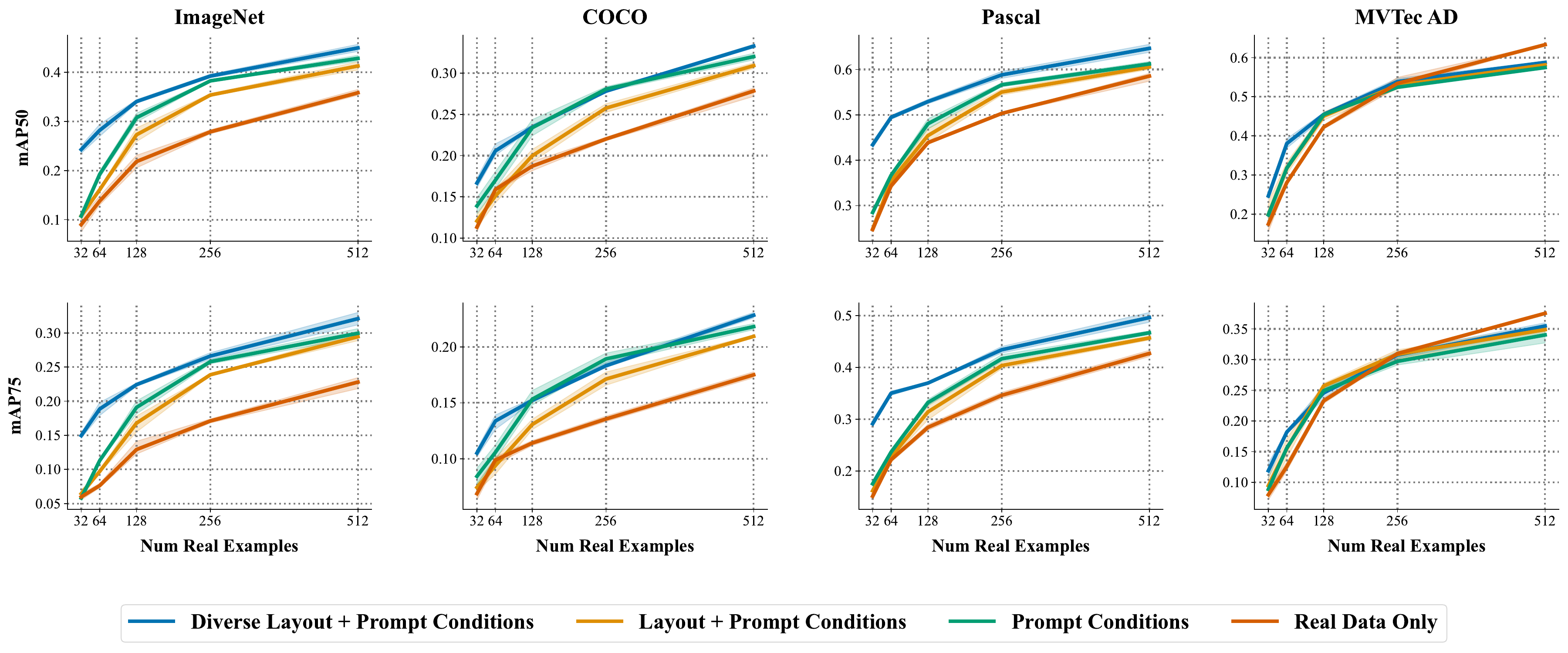}
    \caption{Main synthetic data evaluation. Compares performance across all datasets and all conditioning strategies, as the number of real seed examples per class is scaled. The largest gains come from diverse conditioning, especially in the few-shot regime, where, on average, diverse layout + prompt conditioning improves mAP50 by 84\%.}
    \label{fig:main_results}
\end{figure*}

With a pipeline developed for generating, labeling, and evaluating synthetic data, we can now focus on studying our main experimental question: \textit{what are the key factors that unlock performant synthetic data for object detection?} To answer this question, we first explore how the diversity of conditions for image generation impact results.

\paragraph{Dataset Preparation} We evaluate synthetic data on uniformly four standard object detection tasks spanning a range of domains and difficulties: ImageNet \cite{Imagenet}, COCO \cite{COCO}, Pascal VOC \cite{PascalVOC}, and MVTec Anomaly Detection \cite{MVTecAD}. The inclusion of MVTec tests the generalization of synthetic data to distributions that significantly differ from the training data of current diffusion models \cite{StableDiffusion}. For each dataset, we sample random training sets that contain 32, 64, 128, 256, and 512 real examples. We create three independent trials for each data scale with a different randomization to estimate confidence intervals. On each dataset, the number of object classes varies significantly, ranging from 200 for ImageNet \cite{Imagenet} to 20 for Pascal \cite{PascalVOC}. To ensure that each domain has similar object coverage for a fixed real data scale, we select 20 object classes uniformly at random for each dataset, and restrict images and labels to contain just those objects. Refer to Section~\ref{app:dataset} in the Appendix for more details on the selected object categories for each dataset.

\begin{figure*}
    \centering
    \includegraphics[width=\linewidth]{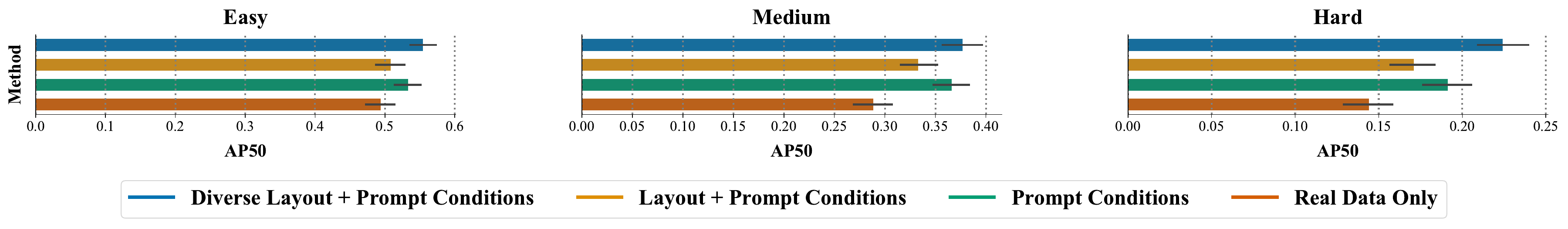}
    \caption{Analysis of synthetic data gains stratified by difficulty. Looking at ImageNet, using 128 real examples as the seed set, we stratify each class into easy, medium, or hard, based off of mAP50 when just training on real examples. While synthetic data helps across the board, the largest synthetic data gains come from the most difficult classes, where we see diverse layout + prompt conditioning helps significantly boosting performance by 57\%.}
    \label{fig:difficulty_results}
\end{figure*}

\paragraph{Modeling Details} We use the YOLOv8 \cite{YOLOv8} architecture from Ultralytics \cite{ultralytics} as our downstream object detection backbone trained on mixtures of real and synthetic data. This model is a single-stage CNN-based object detector commonly used for data-limited tasks due to its efficiency. In our main experiments, we employ the 3.5 million parameter variant of YOLOv8, pretrained on COCO \cite{COCO}. During our fine-tuning, we replace the classification head of the final layer of YOLOv8 with a randomly initialized linear layer, and perform full fine-tuning for 3000 training steps with a batch size of 16, using the AdamW \citep{AdamW} optimizer, a learning rate of \texttt{1e-3}, a weight decay of \texttt{1e-5}, and leave other parameters as PyTorch defaults. We record validation metrics after every epoch of real data, and select the model checkpoint with highest validation mAP50 for reporting results in the main experiments---i.e. early stopping.

\paragraph{Experiment Details} For each dataset, we generate synthetic images and labels as described in Section~\ref{sec:synthetic-generation}, and train detection models following Section~\ref{sec:detection-evaluation}. We employ $\alpha = 0.5$, and $M=10$ for the main experiments, and prepare data as in Paragraph \textbf{Dataset Preparation}, varying the amount of real data observed by downstream models. Equipped with this evaluation protocol, we consider how to identify key factors for generating effective synthetic data. We compare varying synthetic data generation methods to real data baselines that employ a tuned RandAugment \cite{RandAugment}. Synthetic data methods also employ RandAugment, applied to both real and synthetic images. Using RandAugment for all methods is an important choice to avoid multiple covariates that could make results harder to interpret. For all experiments, we report an average across three independent trials, and the 68\% confidence interval. We primarily report mAP50 in our experiments, as this is a popular metric reported in previous works for few-shot object detection.

\subsection{Diversity Is Key}
\label{sec:diversity-is-key}

\paragraph{Setup} To understand the importance of diverse conditions when generating synthetic data, we conduct an experiment scaling the amount of real examples, and training object detection models on mixed datasets of real and synthetic examples. Our goal in this experiment is to understand how \textit{optimally diverse conditions} impact image-first, and labels-first generation strategies akin to Section~\ref{sec:background}. Given that effective methods for labels-first generation of synthetic data have not yet been produced by the field, we aim to simulate this style of generation by conditioning the diffusion model on layouts from real images $c_{\text{layout}}(x_{\text{real}}, y)$ as a proxy. We design four methods, \textit{Real Data Only}, which employs RandAugment on top of real samples when training downstream object detection models, \textit{Prompt Conditions}, which simulates the image-first generation strategy, using prompt conditions $c_{\text{prompt}} (x_{\text{real}}, y)$ produced by a VLM on the few-shot real samples, \textit{Layout + Prompt Conditions}, which simulates the labels-first generation strategy, using  both prompt and layout conditions from the few-shot real samples, and \textit{Diverse Layout + Prompt Conditions}, which provides an optimally diverse set of prompts and layouts from a larger set of real images. This setup now allows us to study how the two paradigms of synthetic data scale as our ability to model the data distribution $p(x_{0}, y)$ improves.

\paragraph{Interpreting The Results} Figure \ref{fig:main_results} compares the performance across all four datasets and conditioning techniques, while scaling the amount of real examples. We find that synthetic data improves model performance on almost all cases. Interestingly, conditioning on just the prompt usually outperforms conditioning on layouts and prompts, suggesting that for weaker models of the data distribution $p(x_{\text{real}}, y)$, image-first generation is preferred. However, as our ability to model $p(x_{\text{real}}, y)$ improves and we have diverse realistic conditions, we observe significant performance gains with layout conditions, especially when we have relatively little real data---at 32 real examples per class, we see a gain of 84\% in mAP. These proxies suggest that as our ability to model the data distribution improves, label-first generation begins to outperform image-first generation of synthetic data. Motivated by these findings, we explore how the task difficulty impacts gains.

\paragraph{Difficulty Impacts Performance} Not all samples are created equally. Some scenarios are harder to learn than others which only increases the importance of generating synthetic data for edge cases. We measure this by stratifying results in Figure~\ref{fig:difficulty_results} into three categories based on the class-wise performance of detection models trained purely on real data. A class is considered easy if its AP50 falls within the bottom third of all classes for that task. It is classified as medium if its AP50 lies within the middle third, and top if its AP50 ranks within the top third of all object classes. We then measure AP50 across all data generation methods for all object classes. Our findings show that as difficult per class increases, performance gains become more pronounced, with synthetic data boosting AP by +13\% for easy objects, to +50\% for hard objects. This indicates that focusing on rare or difficult-to-learn features through synthetic generation yields substantial improvements, pointing to a powerful strategy for enhancing detection models. Future work will explore systematic methods for identifying and synthesizing rare data features.

\subsection{Gains Scale With Diversity}
\label{sec:diversity-scaling}

\begin{figure}
    \centering
    \includegraphics[width=\linewidth]{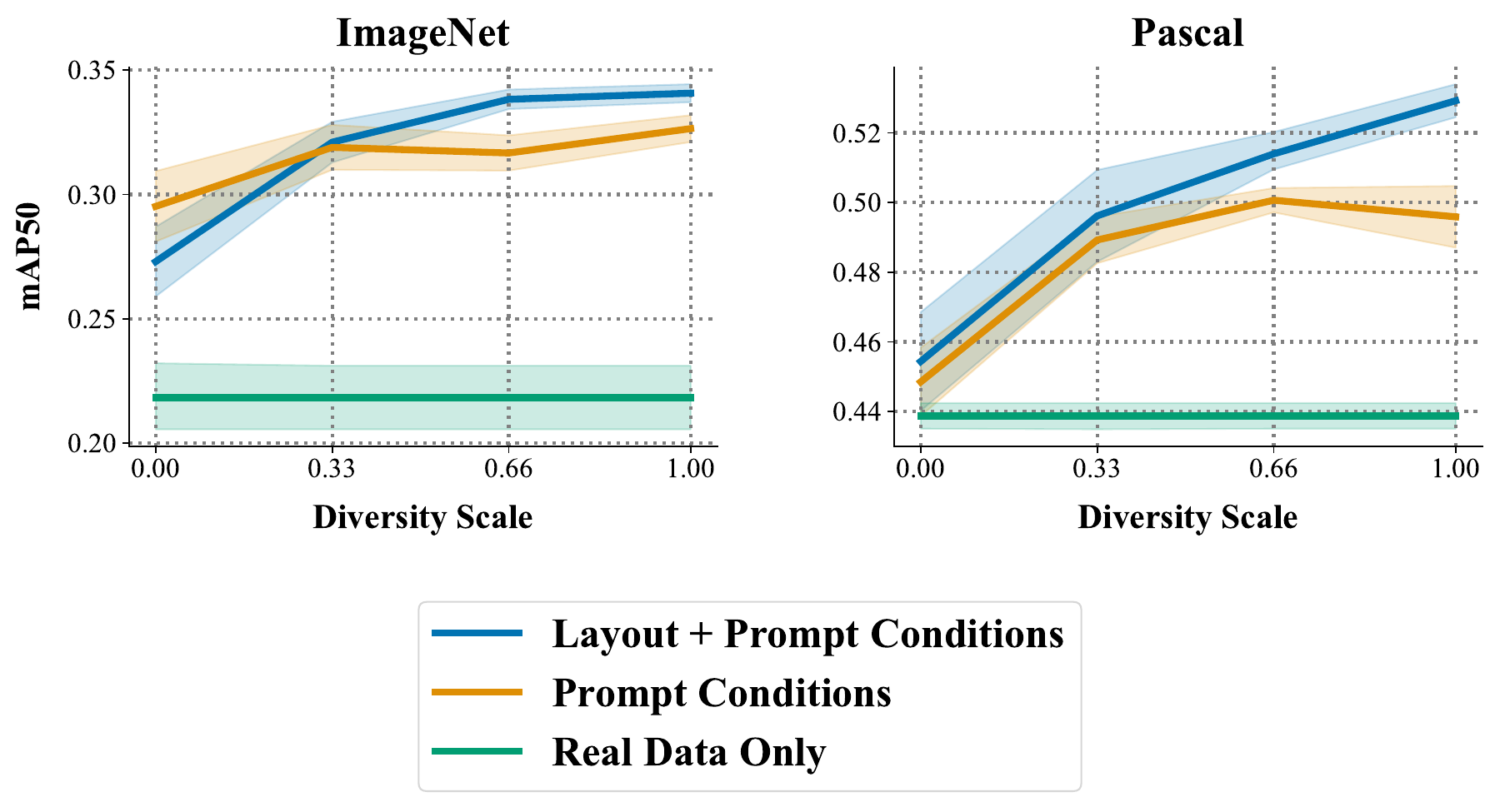}
    \caption{Diversity scaling for prompt and layout conditions. Performance gains scale as diversity increases for Imagenet and Pascal, when starting with 128 real seed samples. Note that the number of synthetic examples stays constant throughout.}
    \label{fig:diversity_scaling_results}
\end{figure}

In previous results, we observe that diverse conditions are a key factor that unlocks effective synthetic data. But given that methods to generate diverse conditions may be far away, an important question is how data generation scales as diversity improves with incrementally better methods---illuminating the path that lies ahead. To simulate this scaling regime, we conduct an ablation that gradually increases the pool of real data from which prompt and layout conditions are drawn, starting from the few-shot real examples, and ending at unique conditions for each synthetic example. We control this sweep via a \textit{diversity scale} hyperparameter where $0.0$ is least diverse, and $1.0$ is most diverse. Figure \ref{fig:diversity_scaling_results} tests detection models trained on data mixtures with gradually more diverse prompt and layout conditions---proxies for image-first and labels-first data generation strategies.

\paragraph{Interpreting The Results} Figure \ref{fig:diversity_scaling_results} examines the gains in diversity by showing how scaling the amount of diversity added affects performance. We see that, for both prompt and layout + prompt conditions, increasing the diversity of the conditions improves performance, but layout + prompt conditions scales better as the conditions improve.

\subsection{Data Scale > Model Size}

\begin{figure}
    \centering
    \includegraphics[width=\linewidth]{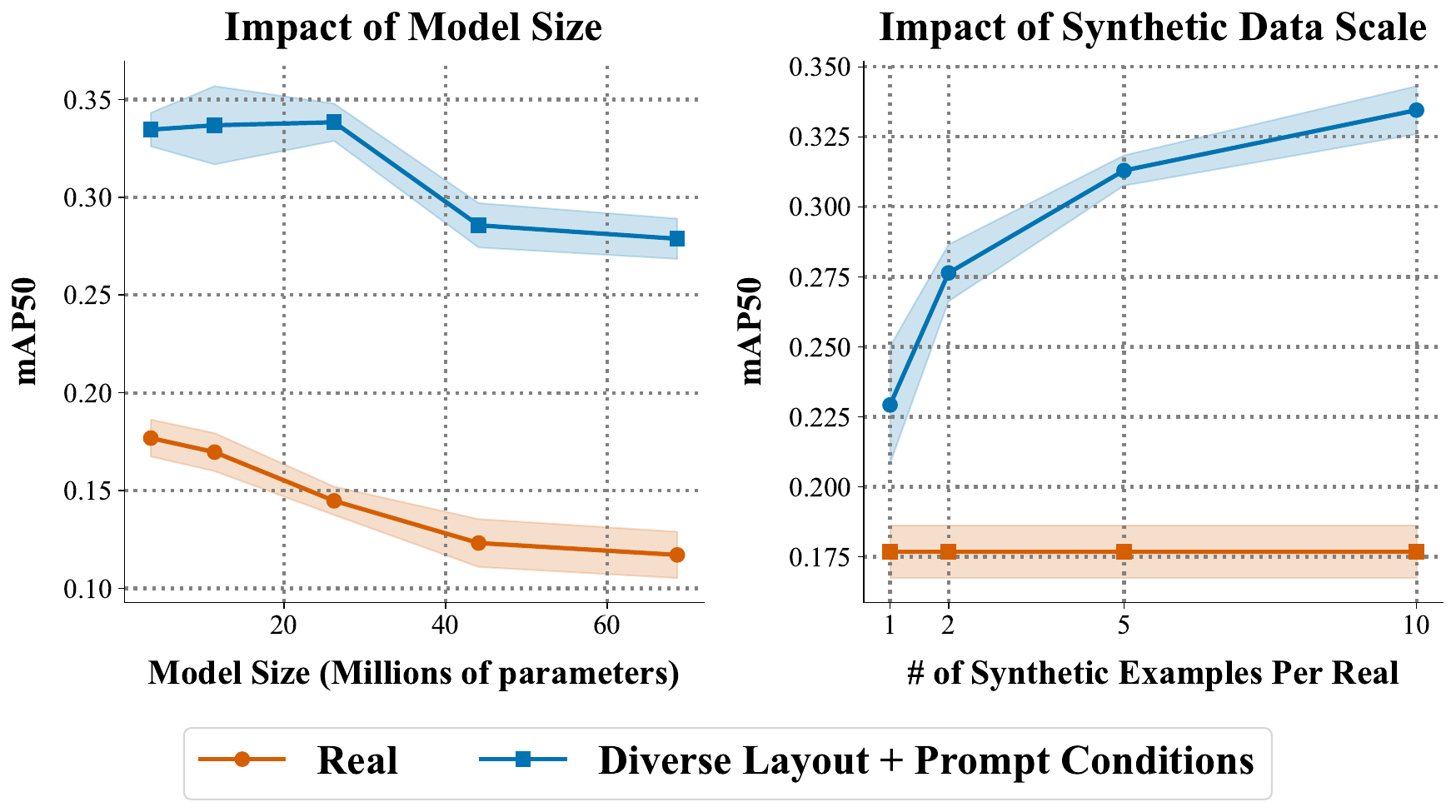}
    \caption{Analysis of model scaling and synthetic-to-real ratio on ImageNet with 128 real examples (left) Performance comparison across YOLOv8 model sizes when training with real data alone versus training with diverse layout and prompt conditioned synthetics. Diverse conditions significantly outperform real examples across all model scales. (right) Impact of varying synthetic-to-real ratios on model performance, demonstrating robust improvements from diverse conditioning across different synthetic-to-real ratios.}
    \label{fig:model_data_scale}
\end{figure}

We next inspect how conditioning scales across different model sizes. Figure~\ref{fig:model_data_scale} compares performance on diverse conditioning compared to real across 5 model sizes, from 3 to 68 million parameters. At all scales, diverse layout and prompt conditioning significantly outperforms just using real examples by at least 2x. Surprisingly, we see performance dip for both settings at the larger model scales, suggesting there may be overfitting to the real examples. Synthetic data allows for training models up to 26 million parameters before we observe overfitting, while models trained on purely real data begin overfitting at just 11 million parameters. In other words, synthetic data allows for training more than a 2x larger model on this task.

We also examine how scaling the amount of synthetic examples generated per real example impacts performance. Figure~\ref{fig:model_data_scale} shows that, as we increase the amount of synthetic data, performance continues to improve, even at a 10-1 synthetic-to-real ratio. 
\section{Discussion}

% Option 1
% Our proposed method for generating labeled synthetic data that leverages VLMs, canny edges, and large-scale diffusion models is a fully end-to-end, automated process that generalizes across domains without making any task-specific assumptions. We show that ultimately what really matters the most to train a reliable vision model is the amount of data available. Diffusion models have unlocked synthetic data for applications where collecting real-world data is not practical. We observe that the more diverse conditions used to guide the generation process, the higher the model performance. Under optimal conditions, we are able to improve model performance on difficult few-shot tasks by up to 177\%. 

% These findings demonstrate the transformative potential of leveraging synthetic data pipelines to overcome data scarcity, redefining how deep learning models are trained!

% Option 2
Our results show that synthetic data from diffusion models can significantly outperform traditional augmentation methods in data-limited settings when optimally controlling the distribution of generated data. We conduct an analysis to explore how different factorizations of the data distribution (i.e. image-first, and labels-first) scale in performance as out ability to model the data distribution improves. Results show that diversity is key when generating synthetic data, and a combination of diverse prompt and layout conditions leads to a relative gain of up to 177\% in mAP over real data across four tasks. We explore the scaling behavior of these gains, and find that as the amount of synthetic data increases, and the diversity of conditions improves, gains produced by our data continue to increase. 

Our pipeline for generating labeled synthetic data is an end-to-end process that generalizes across domains without making any task-specific assumptions. We show that ultimately what really matters the most to train a reliable vision model is the amount of data available. Diffusion models have unlocked synthetic data for applications where collecting real-world data is not practical. We observe that the more diverse conditions used to guide the generation process, the higher the model performance. These findings demonstrate the  potential of synthetic data pipelines to overcome data scarcity, redefining how deep learning models are trained.

%I'm going to take a pass at a draft

{
    \small
    \bibliographystyle{ieeenat_fullname}
    \bibliography{main}
}

\clearpage
\setcounter{page}{1}
\maketitlesupplementary

\section{Dataset Details}
\label{app:dataset}

We employ four datasets in this work: ImageNet \cite{Imagenet}, COCO \cite{COCO}, Pascal \cite{PascalVOC}, and MVTec \cite{MVTecAD}. These datasets are chosen to span a representative set of tasks that researchers and practitioners use when training and evaluating object detection models. The first three of these datasets are object-centric, where entities to detect are one or more such objects present in a scene. The final task, MVTec \cite{MVTecAD}, strongly differs from the first three tasks, and more closely resembles industrial applications. This final task involves detecting defects on various objects, including screws, zippers, pills, and materials like leather, and fabric. This final task is significantly more challenge for pretrained object detection models, as it requires learning to detect parts of objects---often a dent, cut, tear, or the absence of a part.

For each dataset, we create training and validation data for our study by selecting a random set of 20 object classes per dataset, and selecting a random set of $k$ images that contain one or more of those objects. This sampling step is crucial to a balanced evaluation, as directly controlling the amount of training data per object is challenging because images often contain multiple objects, but we want the average \textit{density} of object instances per class per image to match for each dataset. By enforcing that precisely 20 classes are included per dataset, we control the density of labels. Exact classes are listed in Table~\ref{tab:dataset_classes} for reference.

\begin{table*}[]
    \centering
    \begin{tabular}{ll}
        \toprule
        \textbf{Dataset Name} & \textbf{Sampled Classes} \\
        \midrule
        ImageNet & lizard soccer\_ball domestic\_cat motorcycle corkscrew cup\_or\_mug guacamole french\_horn \\
        & elephant lamp harmonica computer\_keyboard racket snail train remote\_control piano pizza porcupine drum \\
        COCO & apple tie train sink surfboard truck chair bench
bottle donut \\
& dog teddy\_bear skis motorcycle scissors
suitcase orange toilet sports\_ball microwave \\
        Pascal & all used \\
        MVTec AD & all used \\
        \bottomrule
    \end{tabular}
    \caption{\textbf{Sampled classes for each dataset.} Note that for Pascal, there are already 20 classes present, and down-sampling is not needed. Similarly for MVTec AD, there are less than 20 classes, so down-sampling is also not needed.}
    \label{tab:dataset_classes}
\end{table*}

In addition to sampling the 20 object instances, we also construct multiple independent data splits of $k$ images. Using three different random seeds, we sample $k \in \{ 32, 64, 128, 256, 512 \}$ real images for each dataset for each independent training data splits. For evaluation, we employ the official validation data for each dataset, and randomly select 1024 images from the validation set that contain instances of the 20 target object classes. In few-shot experiments, we training object detection models on only the $k$ real images in the training data in each split, and we validate on the $1024$ validation images selected from the corresponding validation set, and report confidence intervals.

\section{System Prompts}
\label{app:prompts}

We employ OpenAI's GPT-4o model, queried through the OpenAI API to generate descriptive captions for images in each data---excluding COCO, which already contains such captions in the existing annotations. Given an image from target datasets, we first resize the image to 512 x 512 pixels, then generate a caption for the image using the OpenAI API with the following system prompt jinja template:

\begin{lstlisting}[breaklines=true]
## Instructions For Labeling Images

You are helping us create descriptive prompts and assign labels to images collected from {{ task_description }}.

## Writing Generation Prompts

Help us create image generation prompts for Stable Diffusion XL, a state-of-the-art Diffusion-based text-to-image model. You should write a prompt for generating images like the ones we will show you, but with more details that go beyond the aspects we will ask you to label. Prompts should be 30 words or less.

## Aspects To Label

Help us create accurate labels for the following aspects of the images:
{% for concept_name in concept_names %}
{{ loop.index }}. `{{ concept_name }}`: {{ concept_descriptions[loop.index - 1] }}{% endfor %}

## Options For Labels

For each aspect, select the most accurate label from the following set of options:
{% for concept_name in concept_names %}
{{ loop.index }}. `{{ concept_name }}`: select from {% for option in concept_options[loop.index - 1] %}`{{ option }}`, {% endfor %}{% endfor %}

## Formatting The Response

Format your response in JSON, and adhere to the following schema:
```
{
    "generation prompt": str,{% for concept_name in concept_names %}
    "{{ concept_name }}": str,{% endfor %}
}
```

Ensure that your response is formatted correctly, has a descriptive prompt (refer to section:  Writing Generation Prompts), has accurate labels for each aspect we listed (refer to section:  Aspects To Label), and only uses labels from the set of options we provided (refer to section: Options For Labels).

Thanks for helping us annotate images, do your best to respond accurately.
\end{lstlisting}

The variable \texttt{task\_description} in the template is a description of the target dataset to improve the relevance of captions. For the imagenet dataset, this variable is set to \texttt{"the 2014 ImageNet ILSVRC object detection task"}, for Pascal this is set to \texttt{"the 2012 Pascal VOC dataset"}, and for MVTec this is set to \texttt{"an industrial inspection task, where the goal is to identify defects"}. 

Note as well that we provide optional fields for attribute labels alongside caption generation to improve visual grounding of the captions. These attribute fields are not used for ImageNet, and Pascal, while for MVTec we include a concept \texttt{"has defect"} with a description \texttt{"indicates whether a defect is present"} that takes on values of \texttt{"yes or no"}. We provide only the system prompt, and do not include in-context examples for the caption generation step.

\section{Hyperparameters}
\label{app:hparams}

In this section, we provide the hyperparameter used in synthetic data generation, including parameters of the diffusion process used for image generation, the label generation with Owl-v2 \cite{Minderer2023ScalingOO}, and layout condition extraction. Values for hyperparameters are shown in Table~\ref{tab:hyperparameters}, and code for reproducing our experiments is available at the following anonymous site: \href{https://anonymous-diffusion-augmentation.github.io}{anonymous-diffusion-augmentation.github.io}.

\begin{table*}[]
    \centering
    \begin{tabular}{ll}
        \toprule
        \textbf{Hyperparameter Name} & \textbf{Value} \\
        \midrule
        Number of Independent Trials & $3$ \\
        Confidence Interval Type & $68\%$ \\
        OpenAI Model ID & \texttt{gpt-4o} \\
        SDXL HuggingFace Model ID & \texttt{stabilityai/stable-diffusion-xl-base-1.0} \\
        SDXL VAE HuggingFace Model ID & \texttt{madebyollin/sdxl-vae-fp16-fix} \\
        SDXL ControlNet HuggingFace Model ID & \texttt{diffusers/controlnet-canny-sdxl-1.0} \\
        SDXL ControlNet Conditioning Scale (Diffusers) & $0.5$ \\
        SDXL Image Resolution & $1024$ px \\
        Diffusion Sampling Steps & $20$ \\
        Classifier Free Guidance Weight & $7.5$ \\
        Layout Condition Type & \texttt{canny} \\
        Synthetic Images Per Real $M$ & $10$ \\
        Mixing Ratio $\alpha$ & $0.5$ \\
        Owl-v2 HuggingFace Model ID & \texttt{google/owlv2-base-patch16-ensemble} \\
        Owl-v2 Detection Confidence Threshold & 0.1 \\
        YOLO Model Type & YOLO-v8n (3.5 million parameters) \\
        YOLO Image Resolution & $640$ px \\
        YOLO Batch Size & $16$ \\
        YOLO Optimizer & AdamW \\
        YOLO Learning Rate & $0.001$ \\
        YOLO Weight Decay Rate & $0.00001$ \\
        YOLO Training Steps & $3000$ \\
        \bottomrule
    \end{tabular}
    \caption{Hyperparameters and their values.}
    \label{tab:hyperparameters}
\end{table*}

\end{document}